\documentclass[10pt,twocolumn,letterpaper]{article}

\usepackage{cvpr}              
\usepackage{stfloats}
\usepackage{float}
\usepackage[marginal]{footmisc} 
\usepackage[margin=1in]{geometry}
\usepackage{colortbl}
\usepackage{xcolor} 
\definecolor{best_color}{rgb}{1, 0.7, 0.7}
\definecolor{second_color}{rgb}{1, 0.85, 0.7}
\definecolor{third_color}{rgb}{1, 1, 0.7}
\newcommand{\best}{\cellcolor{best_color}}
\newcommand{\second}{\cellcolor{second_color}}
\newcommand{\third}{\cellcolor{third_color}}
\definecolor{cvprblue}{rgb}{0.21,0.49,0.74}
\usepackage[pagebackref,breaklinks,colorlinks,allcolors=cvprblue]{hyperref}

\title{Advancing high-fidelity 3D and Texture Generation with 2.5D latents}

\author{
\href{mailto:xyangbk@connect.ust.hk}{\textcolor{black}{Xin Yang}}$^{1,2*}$ \href{mailto:jlin695@connect.hkust-gz.edu.cn}{\textcolor{black}{Jiantao Lin}}$^{1*}$ Yingjie Xu$^1$ Haodong Li$^1$ \href{mailto:yingcongchen@hkust-gz.edu.cn}{\textcolor{black}{Ying-Cong Chen}}$^{1,2\dagger}$ \\
$^1$HKUST(GZ) $^2$HKUST
}

\begin{document}

\twocolumn[{
\maketitle
\begin{figure}[H]
    \vspace{-2em}
    \hsize=\textwidth
    \centering
    \includegraphics[width=2.2\linewidth]{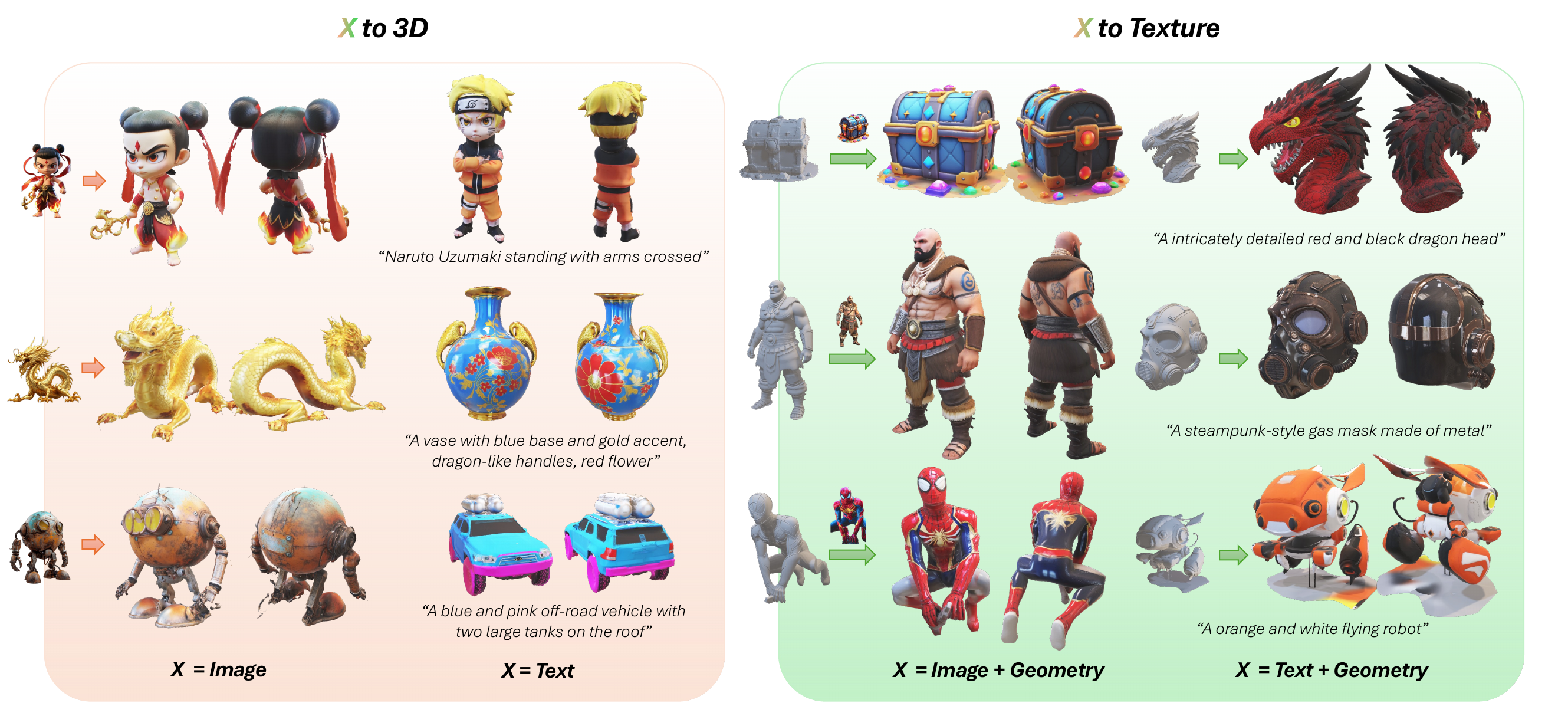}
    \vspace{-0.5em}
    \caption{\textbf{Examples of ``X'' to 3D and ``X'' to texture generation.} In this paper, we propose a new approach to bridge 3D generation with 2D diffusion priors, i.e., the generation of 2.5D latent. By leveraging the advantage of 3D-like representation and the prior of pretrained 2D diffusion models to the best extent, we achieve not only 3D generation with high-fidelity, but also excels in geometry-conditioned texture generation with image or text conditions. Please zoom in for details.}
    \label{fig:teaser}
\end{figure}
}]

\noindent\footnotetext{* Equal contribution \\ $\dagger$ Corresponding author}


\begin{abstract}
Despite the availability of large-scale 3D datasets and advancements in 3D generative models, the complexity and uneven quality of 3D geometry and texture data continue to hinder the performance of 3D generation techniques. In most existing approaches, 3D geometry and texture are generated in separate stages using different models and non-unified representations, frequently leading to unsatisfactory coherence between geometry and texture.
To address these challenges, we propose a novel framework for joint generation of 3D geometry and texture. Specifically, we focus in generate a versatile 2.5D representations that can be seamlessly transformed between 2D and 3D. 
Our approach begins by integrating multiview RGB, normal, and coordinate images into a unified representation, termed as 2.5D latents. 
Next, we adapt pre-trained 2D foundation models for high-fidelity 2.5D generation, utilizing both text and image conditions. 
Finally, we introduce a lightweight 2.5D-to-3D refiner-decoder framework that efficiently generates detailed 3D representations from 2.5D images.
Extensive experiments demonstrate that our model not only excels in generating high-quality 3D objects with coherent structure and color from text and image inputs but also significantly outperforms existing methods in geometry-conditioned texture generation.
The code and model of this paper is available at \href{https://github.com/AbnerVictor/High-fidelity-2.5D-Generation}{https://github.com/AbnerVictor/High-fidelity-2.5D-Generation}.
\end{abstract}
\section{Introduction}
\label{sec:intro}
Recently, 3D generative models have been advancing rapidly, largely due to the emergence of large-scale 3D datasets. However, the complex nature of 3D data, which encompasses both geometry and texture information, presents challenges in curating datasets with high-quality ground truths in both domains. This limitation hinders the advancement of high-fidelity 3D generation. In particular, achieving high-quality textures in 3D data is often more difficult, leading to an imbalanced development between geometry and texture generation.

For example, we observe that most state-of-the-art 3D generation models~\cite{li2025triposg, hunyuan3d22025tencent, zhang2024clay, li2024craftsman} adopt a two-stage 3D generation pipeline, consisting of a geometry generation stage followed by a texture generation stage. 
During the texture generation stage, one must exert additional efforts~\cite{huang2024mvadapter, xiong2024texgaussian, zheng2024mvd} to ensure that the generated texture aligns with the pre-generated geometry. 

A recent approach, TRELLIS~\cite{xiang2024structured}, seeks to integrate geometry and texture generation through a 3D structured latent (SLat) generation pipeline. Although TRELLIS is capable of producing highly aligned 3D assets with refined structures, we find that the texture quality of its outputs is often inferior to that of multi-view-based 3D generative models~\cite{lin2025kiss3dgen, huang2024mvadapter, wu2024unique3d}, which utilize 2D generative priors.

Motivated by this observation, we propose a straightforward solution: by combining the advantages of the 3D structured latent representation with 2D generative priors, we can achieve joint generation of both geometry and texture with high quality. 

\begin{figure}
    \centering
    \includegraphics[width=1.0\linewidth]{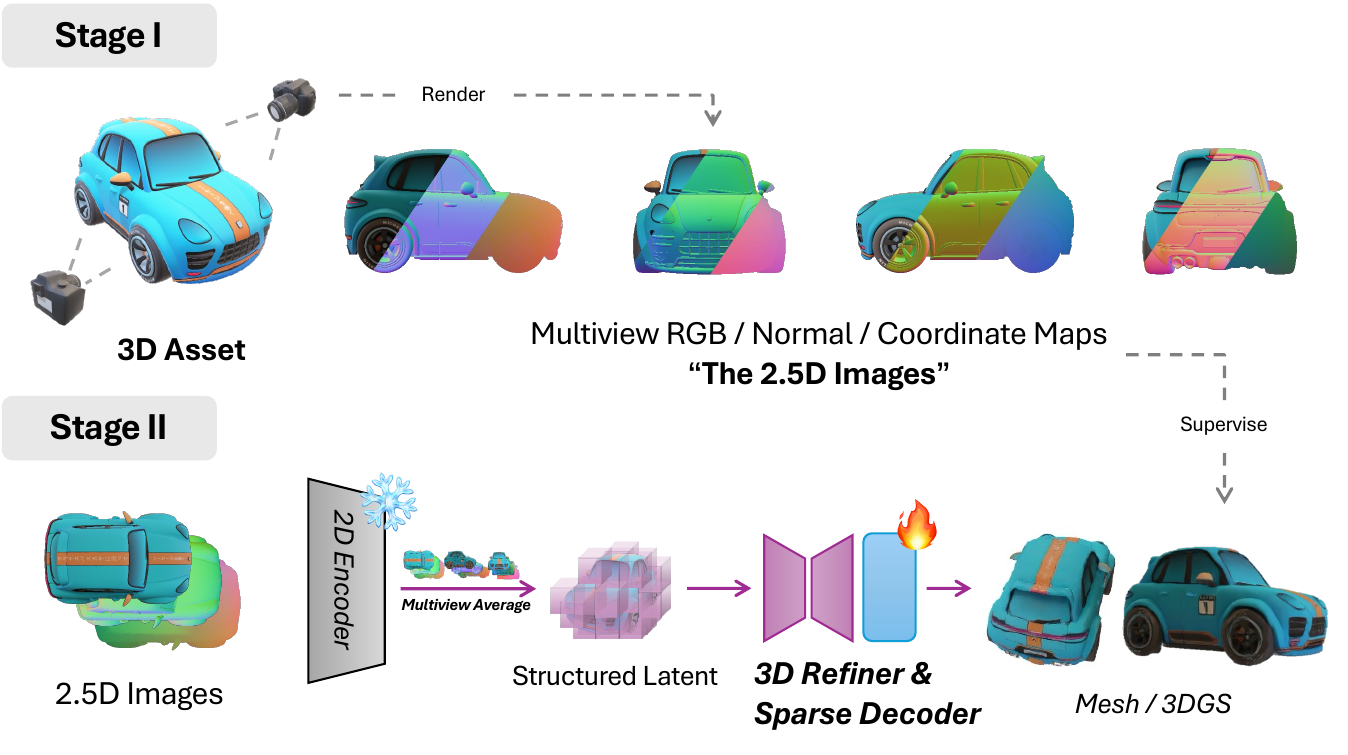}
    \vspace{-2em}
    \caption{\textbf{The multiview 2.5D representation.} In this paper, we curate a 2.5D dataset by rendering the multiview RGB, normal and coordinate maps from 3D assets. 
    With the dataset, we encode the image of each modalities into 2D latents with 2D VAE, and project the latents into 3D voxel space with multiview averaging. Then, we train our 3D refiner and decoder to reconstruct the 3D assets (3D gaussian splats or mesh) from the aggregated structured latents.}
    \label{fig:2-5D-image}
\end{figure}

In this paper, we introduce a novel 3D generative framework based on a 2.5D latent representation which designed to bridge the gap between 2D image-like latents and 3D structured latents. Concretely, a 3D object typically encompasses texture, surface, and geometry information, which can be effectively captured by multiview RGB, normal, and coordinate images, as illustrated in Fig.~\ref{fig:2-5D-image}.

These multiview images in necessary modalities forms the 2.5D representation of the object. With the coordinate components of the 2.5D representation, we can easily project the multiview features into the structured 3D voxel space, which enables 3D native asset reconstruction. Compared with the native 3D representations such as voxel grid and triplane, the 2.5D representation is essentially images, and thus much easier to process. 
Hence, we can generate such 2.5D latents by adapting 2D multiview generative models as previous works~\cite{lin2025kiss3dgen,wu2024unique3d,long2023wonder3d}.

To facilitate this goal, we first present the 3D refiner-decoder modules to reconstruct 3D Gaussian Splats and mesh from the 2.5D latent. Second, we propose a novel Mixture-of-LoRA architecture, which is tailored for multi-modality 2.5D latent generation. 
Additionally, by reusing the pretrained Mixture-of-LoRA adapters, we further fine-tune our framework to enable the partial generation of 2.5D latents, specifically for high-fidelity geometry-conditioned texture generation. This approach significantly outperforms existing methods in both image- and text-conditioned texture generation.

In our experiments, the 2.5D generation framework demonstrate a significant advantage in generating high-quality, geometry-texture coherent outcomes for both 3D and texture generation tasks compared to state-of-the-art approaches. We summarize the main contributions as follows:
\vspace{-0.5em}
\begin{enumerate}
    \item We propose a novel 2.5D latent generation framework for image-or-text-to-3D generation and geometry-conditioned texture generation. The framework bridges the native 3D generative approach with multiview generative models and produces high quality generation results.
    \item We propose the novel 2.5D refiner and decoder model to reconstruct the 3D assets, then optimize the 3D assets with multiview guidance to achieve best visual fidelity. 
    \item We introduce a novel Mixture-of-LoRA approach to adapt pretrained 2D generative models for 2.5D latent generation, which improves the multi-modal generation quality.
\end{enumerate}

\section{Related Works}
\label{sec:related_works}
\subsection{2D Diffusion Models}
Diffusion models~\cite{SohlDickstein2015DeepUL,ho2020denoising} define a Markov process that gradually transforms data into noise through a forward diffusion process and then learns to reverse this transformation to generate data from noise. They have demonstrated remarkable performance in image~\cite{rombach2022high,he2024lotus,dalle,imagen,he2024disenvisioner} and video generation~\cite{ho2022imagen,blattmann2023stable,blattmann2023align,ho2022video}. However, conventional diffusion trajectories often exhibit significant curvature~\cite{Karras2022ElucidatingTD,liu2022flow}, which increases the number of required solver steps and amplifies error accumulation. To address this, Rectified Flow~\cite{liu2022flow,albergo2022building,lipman2023flow} has been proposed, enforcing a straight-line trajectory between noise and data. Flow-based diffusion models have shown strong empirical performance and high efficiency, as demonstrated by recent state-of-the-art systems such as Stable Diffusion V3\cite{esser2024scaling} and Flux.1~\cite{blackforest2024flux}. In this work, we adopt Flux.1 as the backbone diffusion model for 2.5D generation, leveraging its rectified flow design to enable high-quality and efficient image synthesis, which serves as a strong prior in our 3D reconstruction pipeline.

\subsection{3D Generative Models}
\subsubsection{Multiview Diffusion Models}
Multiview diffusion models~\cite{zheng2024mvd,lu2024direct2,miao2024dsplats,qin2025distilling,chen20243d,lin2025kiss3dgen,shi2023mvdream,xu2024flexgen} have focused on synthesizing geometrically consistent object views from varying camera poses, conditioned on image or text prompts. Early efforts in multiview generation models, such as MVDream~\cite{shi2023mvdream}, have successfully adapted pre-trained text-to-image diffusion
models to generate object-centric multiview images. To enhance the geometric fidelity of the generated 3D content, methods such as Direct2.5~\cite{lu2024direct2} and Kiss3DGen~\cite{lin2025kiss3dgen} incorporate geometry-aware designs and normal-guided supervision, producing multiview normal maps that facilitate high-quality 3D asset reconstruction. Other approaches leverage explicit 3D representations to improve realism and structural integrity. ~\citet{miao2024dsplats,qin2025distilling} distill multiview diffusion outputs into compact 3D generators using Gaussian splatting~\cite{kerbl20233d}, enabling efficient single view 3D reconstruction with enhanced visual quality. In addition, auxiliary modules have been proposed to improve cross-view consistency and 3D reasoning. ~\citet{chen20243d} introduces a modular architecture called 3D-Adapter that injects 3D priors into diffusion pipelines, while ~\citet{zheng2024mvd} constructs a volumetric feature representation by aggregating multiview image features, leading to more accurate geometry reconstruction.

\subsubsection{3D Diffusion Models}
In contrast to multiview diffusion models that rely on synthesizing consistent 2D projections for 3D reconstruction, 3D diffusion models aim to directly recover complete 3D structures by progressively denoising from random noise in native 3D representations such as point clouds, voxels, or meshes. These models bypass intermediate image synthesis and focus on high-fidelity 3D reconstruction or generation.

~\citet{xiang2024structured} introduces a unified structured latent space for 3D content (SLAT), supporting diverse representations like meshes, Gaussians, and NeRF. ~\citet{zhang2024clay} uses a DiT-based latent diffusion model to generate neural surface representations and materials. It combines a multi-resolution VAE with PBR-aware training and supports conditional generation from text or images. ~\citet{li2024craftsman} performs native 3D diffusion directly in volumetric space to generate coarse mesh shapes. It integrates a normal-guided surface refiner that enhances geometry. ~\citet{liu2023meshdiffusion} directly applies score-based diffusion to mesh vertex coordinates and face connectivity. It models 3D meshes in a native representation, enabling faithful and topology-preserving mesh generation without intermediate voxelization or rendering supervision.

\begin{figure*}[htbp]
    \centering
    \includegraphics[width=1.0\linewidth]{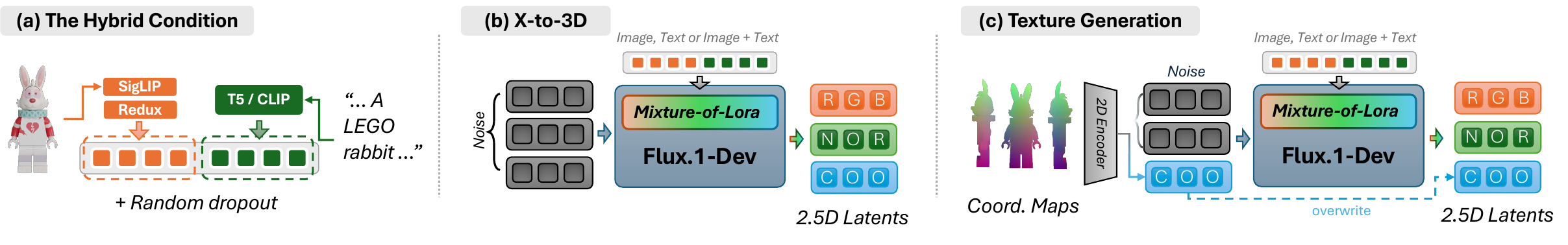}
    \vspace{-2.5em}
    \caption{\textbf{Our proposed framework.} In this paper, we propose a unified framework for (b) text or image to 3D generation and (c) geometry-conditioned texture generation. To provide the (a) hybrid image-text condition for 3D or texture generation, we adopt the off-the-shelf image encoder SigLIP~\cite{zhai2023sigmoidlosslanguageimage} and Flux.1-Redux~\cite{blackforest2024flux} image embedder with the T5 and CLIP text encoder. During the training, we \textbf{randomly dropout the image or text condition} to maintain the model's ability to generate coherent content with "X`` conditions and achieve better model performance, please refer to our experiment section for more details.}
    \label{fig:framework}
\end{figure*}

\section{Method}
\label{sec:method}
In this section, we provide more details about our framework. In Sec.~\ref{sec:2_5D_latent}, we present a simple yet effective approach to composing multiview RGB, normal, and coordinate images into the 2.5D latent. Then, we introduce a refiner-decoder architecture to translate the 2.5D representation into explicit 3D representations such as 3DGS and mesh. In Sec.~\ref{sec:multi-lora}, we introduce a novel solution of LoRA fine-tuning the 2D diffusion model for 2.5D image generation. 

\subsection{Bridging 3D and 2D with 2.5D Representation}
\label{sec:2_5D_latent}

In our framework, we utilize an image-like 2.5D representation (Fig.~\ref{fig:2-5D-image}, stage I) for object encoding and decoding. Unlike 3D representations, the 2.5D representations are essentially images, however, it combines the multiview RGB, normal and coordinate images to represent the volume, surface and texture information that are sufficient for high-quality 3D objects reconstruction. Although some previous works~\cite{lin2025kiss3dgen,wu2024unique3d,long2023wonder3d,wang2024crm} have explored the potential of similar representations that composed of rgb with either coordinate or normal map, the absence of either normal or coordinate map in their framework could introduce extra difficulties in the 3D reconstruction stage.

\noindent \textbf{From 2.5D to 3D.} 
While most previous works adapted sparse-view reconstruction approaches to reconstruct the object from either rgb, normal or coordinate information, their result often limited to specific 3D representation, and sometimes low-quality in the unseen areas of the multi-view results. 
Inspired by the 3D structured latent representation proposed by Trellis~\cite{xiang2024structured}, we also create a structure latent representation, which can then be decoded to various 3D representations in high-quality.
For pipeline simplicity, we choose the VAE of the 2D diffusion model (i.e. Flux.1-dev) to encode the 2.5D images into 2D latents. Then, we project the 2D latents with $C$ channels and $N$ views $x_{2d} \in R^{N \times 3 \times C \times H \times W}$ into a 3D latent voxel grid $x_{3d} \in R^{3C \times X \times Y \times Z}$ accordingly. As shown in stage II of Fig.~\ref{fig:2-5D-image}, the projection of the 2D latent introduces a voxel-like 3D latent representation that contains the color, surface and coordinate information.
Following~\cite{xiang2024structured}, we aim to introduce a sparse 3D decoder to reconstruct the structured latent into explicit 3D representations such as 3D gaussian splats or mesh.

\noindent \textbf{The 3D Refiner and Decoder.} In practice, the occluded areas in the multiview-to-3D projection could leads to undesired holes during the reconstruction. Hence, besides the decoder, we further involve a 3D UNet to refine the initial structured latent. 
Specifically, we train the refiner model to predicts both the refined structured latent $\hat{x}_{3d}$ with its occupancy $\hat{x}_{occ} \in R^{X \times Y \times Z}$. 
Notably, we init $x_{occ}$ by identifying non-zero values in $x_{3d}$, then predicts the refined 3D feature $\hat{x}_{3d}$ with an occupancy bias $b_{occ}$ with the refiner. The refined occupancy $\hat{x}_{occ}$ is defined by:
\begin{equation}
\label{eqn:OCC}
\hat{x}_{occ} = \begin{cases}
1, &\text{if}~x_{occ} + b_{occ} > 0.5\\
0. &\text{otherwise}
\end{cases}
\end{equation}

Then, we train the decoder to convert the refined structure latent into target 3D representation such as 3D gaussian splatting or mesh. Following ~\cite{xiang2024structured}, we adopt the sparse transformer for decoder, and train the refiner unet and 3DGS decoder in an end-to-end manner from scratch with rendering loss only. For the mesh decoding, we fixed the pretrained refiner and train the mesh decoder only. Please refer to our appendix for more details.

%

\subsection{Model Fine-tuning Via Mixture-of-LoRA}
\label{sec:multi-lora}
Inspired by~\citet{lin2025kiss3dgen}, we aims to finetune a pretrained 2D diffusion model, i.e. Flux.1-dev~\cite{blackforest2024flux}, via Low-Rank Adaptation (LoRA)~\cite{hu2022lora} for 2.5D image generation.
As our 2.5D image comprise RGB, normal, and coordinate images, extending the 2D diffusion model to the joint generation of multi-modal images using LoRA adapters with limited data presents a challenge. 
Furthermore, we observe that simply finetune a single LoRA adapter with large rank (e.g. 384) do not achieves the best performance, both quantitatively and qualitatively (see Sec.~\ref{sec:lora-vs-mol}).
\begin{figure}
    \centering
    \includegraphics[width=1.0\linewidth]{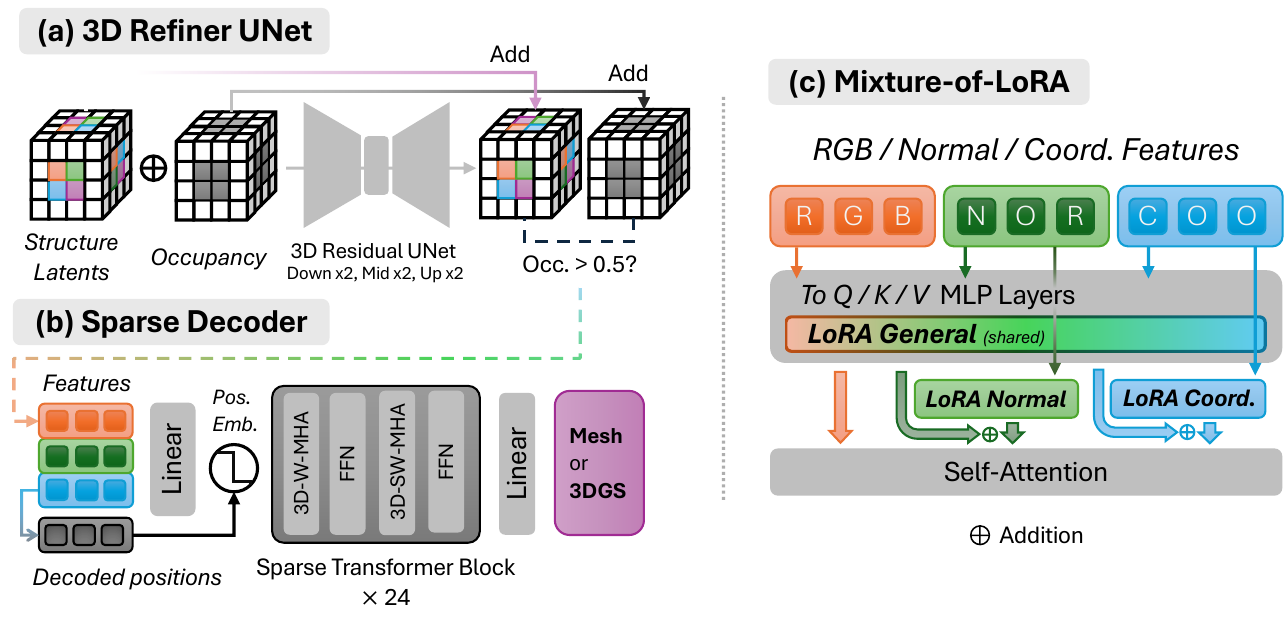}
    \vspace{-2em}
    \caption{\textbf{The architecture of proposed modules.} In order to fix the occluded areas in the structure latent aggregated from our 2.5D latent, we introduce the \textbf{(a)} 3D Residual UNet to refine the feature and occupancy field. Then, we apply the \textbf{(b)} sparse transformer decoder~\cite{xiang2024structured} for 3DGS or mesh reconstruction. \textbf{(c)} For each MLP block, we introduce three separate LoRA adapters tailored for 2.5D latent generation. Specifically, while the general LoRA layer is shared by all modalities, we feed the normal and coordinate (coord.) features into the auxiliary LoRA layers for extra feature projection, then we merge the results of the general and normal or coord. LoRAs via a simple addition.}
    \vspace{-2.0em}
    \label{fig:arch-all}
\end{figure}
Instead, in this paper, we propose a novel scheme of LoRA fine-tuning for multimodal image generation which we refer to as Mixture-of-LoRA (MoL) for simplicity. Instead of applying a single LoRA adapter to the base model, we introduce separate LoRA layers for multi-modal generation, as shown in Fig.~\ref{fig:arch-all}(c).

Concretely, in each MLP layer $F$ with LoRA adapters, we split the latent $x \in R^{B\times L \times C}$ into RGB, normal and coordinate parts $x_{i}$, where $i \in \{\textit{rgb}, \textit{normal},\textit{coord}\}$. Likewise, we implement three LoRA adapters $\hat{F}_{j}$ and $j \in \{\textit{general}, \textit{normal}, \textit{coord}\}$. Denoted the projected feature from $x_i$ as $h_i$, we produce $h_i$ following:
\begin{equation}
\label{eqn:MoL}
h_{i} = \begin{cases}
F(x_i) + \hat{F}_{\text{general}}(x_{i}), &\text{if}~i = \text{``rgb''}\\
F(x_i) + \hat{F}_{\text{general}}(x_{i}) + \hat{F}_{i}(x_{i}). &\text{otherwise}
\end{cases}
\end{equation}


In practice, we apply lower ranks (e.g., 64) for the ``general'' adapter to better preserve the RGB prior of the base model, and use higher ranks (e.g., 128) for the adapter of the other modalities. Notably, we enable the ``general'' adapter to learn both RGB and cross-modal information, which we find important in our experiments.

Compare to the common usage of LoRA adapter, which denoted as $h = F(x) + \hat{F}_{\text{general}}(x)$, our approach leads to significantly improved performance in 2.5D image generation.
Please refer to Sec.\ref{sec:lora-vs-mol} for more details.

\noindent \textbf{Extended Rotary Positional Embedding.}
To adapt the pretrained models to our novel requirements, we preserve the intrinsically integrated rotary positional embedding (RoPE) within these models while introducing a straightforward modification. Specifically, we consider each modality in the 2.5D latent space as an independent 2D latent space for spatial RoPE. We then add constant biases to an auxiliary dimension for cross-modal feature differentiation. Please refer to our appendix for implementation details.

\noindent \textbf{Joint training of text and image condition.}
Unlike previous works, we build a versatile framework for text and image conditioned generation of the 2.5D latent. As we chose the pretrained 2D diffusion model (i.e., Flux-dev.1) as the base model for 2.5D generation, we adopt the off-the-shelf text encoder (T5 and CLIP) and image encoder (SigLIP~\cite{zhai2023sigmoidlosslanguageimage} image encoder with Flux-Redux~\cite{blackforest2024flux} image embedder) to embed the text and image conditions respectively.
During the model finetuning, we randomly drop the image or text condition, which we found beneficial to both text and image conditioned generation tasks. Please refer to our experiment section for more details.



\section{Experiment}
\subsection{Training details}
For our base model for X-to-3D generation, we train the General, Normal and Coord. LoRA adapters with the rank of 128 on Flux.1-Dev model. 
We train the LoRA adapters using a learning rate of 5e-5 for 100k steps with batch size of 8 on 8 Nvidia A800 GPUs. During the training, we randomly dropout the image condition by 80\%, and randomly dropout the text condition while image condition is provided by 50\%. Besides, we also dropout both text and image conditions by 8\%.
For our texture generation model, we finetune the LoRA adapters from the pretained X-to-3D model for another 20k steps, with a decreased dropout rate of 50\% for image condition. 

For the 3D refiner UNet and the sparse decoder of 3D Gaussians, we we adopt a voxel resolution of $64^3$ for the structured latent, and train the models use a learning rate of 2e-4 with batch size of 4 for 150k steps. 
Then, we froze the 3D refiner UNet and continue to train the sparse decoder for mesh decoding.

\subsection{Dataset}
Due to the heterogeneous quality of the original Objaverse dataset, we initially excluded assets that lacked texture maps or exhibited low mesh fidelity. A comprehensive manual curation process was then performed to remove low-quality samples, including incomplete geometries, scanned planar surfaces, and large-scale scene compositions. Additionally, to address inconsistencies in object orientations, we manually annotated the canonical front-facing direction for each object. Then, we further apply a filtering following Trellis~\cite{xiang2024structured} to remove the items with low aesthetic score. 

In parallel, we collected a separate dataset consisting of 4,000 high-quality 3D cartoon-style human figures from online sources, specifically aimed at supporting stylized character generation tasks. By combing the filtered data from Objaversa with items selected form the human figure dataset, we resulted in a dataset of 26,000 high-quality 3D objects.

For data rendering, we used Blender with a fixed camera distance of 4.5 units and a field of view (FoV) of 30 degrees. Each object was rendered from six distinct views to construct our 2.5D representation: four views were captured at 90-degree intervals in azimuth with a fixed elevation of 5 degrees—starting from the annotated front-facing direction—and an additional top-down view was rendered from a 90-degree elevation aligned with the frontal axis. For each view, we rendered RGB images, surface normal maps, and coordinate maps at a resolution of 512×512 pixels. The RGB multi-view images were then used as inputs for generating 3D-aware textual annotations via GPT-4V and Florence2~\cite{xiao2023florence}.

\subsection{Evaluation}
For evaluation, we conducted both quantitative and qualitative comparisons. Quantitatively, we used the Google Scanned Objects (GSO) dataset~\cite{downs2022google}. For the text-to-3D and text-to-texture task, we randomly sampled 100 objects and generated four-view renderings with GPT-4V annotations as text prompts. For image-to-3D and image-to-texture, we sampled 200 objects and rendered a single front-facing view as input for evaluation.

Additionally, we collected 3D models with complex geometry and textures from public sources for qualitative comparison, enabling visual assessment of reconstruction quality and texture fidelity.

\subsubsection{Metrics}
We employ a comprehensive set of quantitative metrics to evaluate both the visual fidelity and geometric accuracy of the generated 3D models. 
For the text-to-3D and text-to-texture task, where precise 3D ground truth is unavailable, we assess semantic alignment using the CLIP score~\cite{hessel2021clipscore} by rendering four orthogonal views of each generated object. In addition, we use Q-Align~\cite{wu2023qalign}, a large multi-modal evaluation model, to assess the quality and aesthetic consistency of the rendered outputs.

For the image-to-3D and image-to-texture task, evaluation is conducted from both 2D and 3D perspectives. The 2D visual quality is measured by rendering novel views from the generated meshes and comparing them against ground truth images using PSNR, SSIM, and LPIPS. To assess 3D geometric quality, we align the coordinate systems of the generated and reference meshes, and normalize them within a unit cube \( [-1, 1]^3 \). We then uniformly sample 16{,}000 points from the mesh surfaces and compute Chamfer Distance (CD) and F-score (FS) at a threshold of 0.1 to quantify geometric similarity.


\subsection{Comparison with State-of-the-Art Methods}
In this section, we compare our approach with several state-of-the-art methods across three tasks: text-to-3D generation, image-to-3D generation and texture generation.
For text-to-3D generation, we compared our method with four representative baselines: the diffusion-based model CraftsMan~\cite{li2024craftsman} and Trellis~\cite{xiang2024structured}, Hunyuan3D-1~\cite{yang2024hunyuan3d} which builds on a large reconstruction model (LRM), Unique3D~\cite{wu2024unique3d}, a two stage generation method.
For image-to-3D generation, We compared our method against three baselines: 
the latent diffusion model 3DTopia~\cite{hong20243dtopia}, Hunyuan3D-1~\cite{yang2024hunyuan3d}, and Trellis~\cite{xiang2024structured}.
For texture generation, we compared our method against Hunyuan2.0~\cite{hunyuan3d22025tencent}, MV-Adapter~\cite{huang2024mvadapter} which generates a single image from text, and then expand it into six multi-view images in image-to-texture generation, and SyncMVD~\cite{liu2024text} which generates a central image from text and predicts synchronized multi-view images via a cross-view fusion module, TexGaussian~\cite{xiong2024texgaussian} using 3D Gaussians in text-to-texture generation.

\begin{figure}
    \centering
    \includegraphics[width=1.0\linewidth]{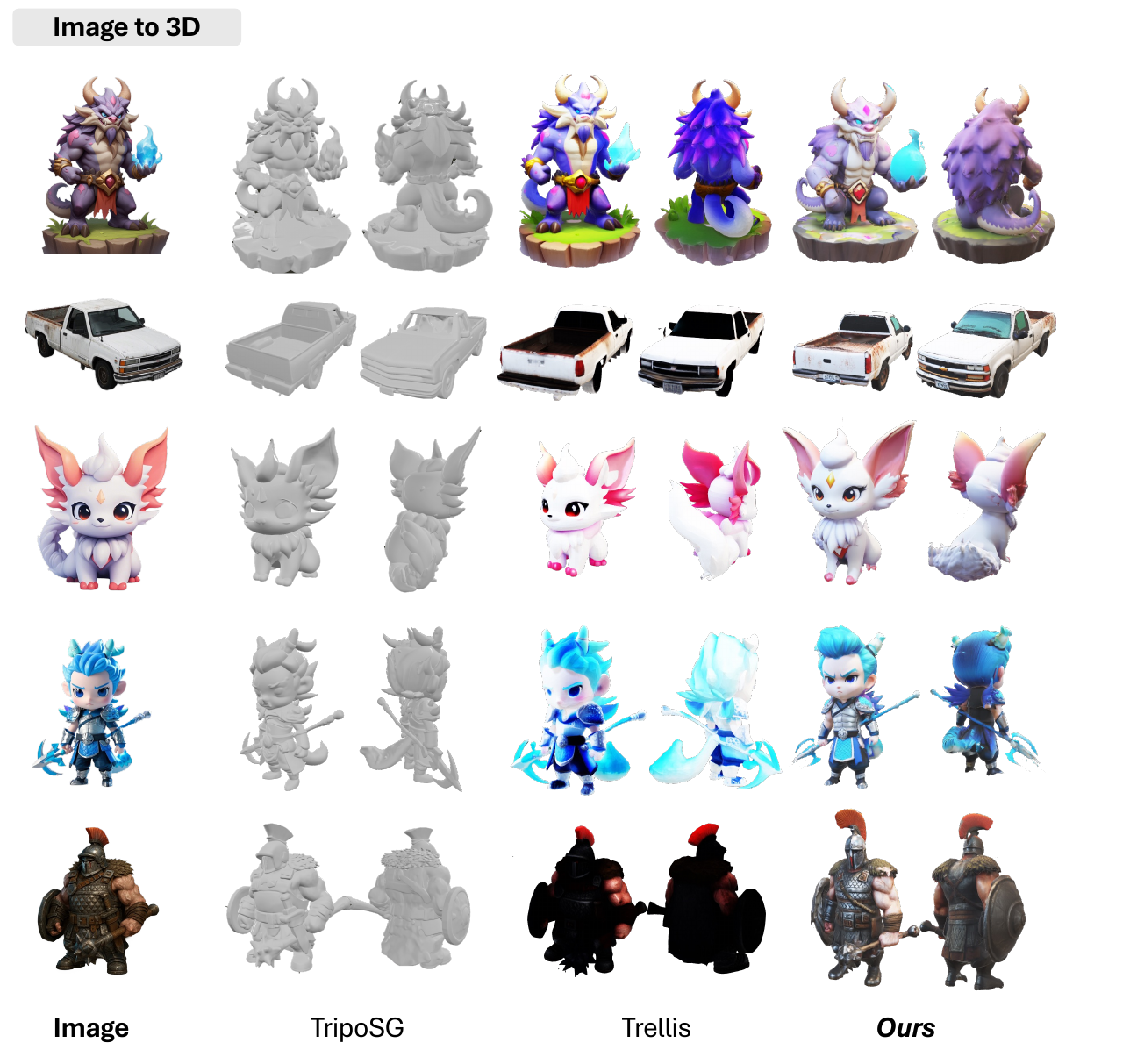}
    \vspace{-2em}
    \caption{Qualitative comparison on \textbf{image-to-3D generation}. Please zoom in for detail.}
    \label{fig:img-to-3D}
\end{figure}

\noindent\textbf{Image-to-3D generation.}
 As shown in Table~\ref{tab:quantitative_comparison}, our method achieves competitive performance across both geometric and perceptual metrics. Notably, our results are comparable to Trellis, despite Trellis relying on significantly larger-scale training data and a much larger model. In contrast, our method is designed to be more data- and compute-efficient, yet achieves the highest PSNR and a competitive LPIPS score, ranking closely behind Trellis. Our SSIM score is also comparable to top-performing methods, indicating strong perceptual consistency and visual fidelity in the rendered views. These results suggest that our approach effectively reconstructs high-quality 3D shapes from single-view inputs, achieving a favorable trade-off between visual quality and computational cost.
 In addition to the quantitative results, Figure~\ref{fig:img-to-3D} presents a qualitative comparison under the image-to-3D setting. Compared to TripoSG~\cite{li2025triposg} and Trellis, our method produces significantly more faithful and detailed textures. For example, our results better preserve color contrast, facial features, and fine-grained patterns such as fur, eyes, and accessories. Furthermore, our textures remain consistent across multiple viewpoints, reflecting better integration between appearance and structure.
\begin{table}[h]
\centering
\resizebox{1.0\linewidth}{!}{
\begin{tabular}{ccccccc}
\toprule
Method   & Dataset Size    & CD$\downarrow$   & FS$\uparrow$   & PSNR$\uparrow$  & SSIM$\uparrow$   & LPIPS$\downarrow$ \\
\midrule
CraftsMan & 170K              & 0.178            & 0.739          & --              & --               & --             \\
Hunyuan3D-1   & N/A           &\third{0.153}     & \third{0.768}  & 17.154          & 0.868            &  0.154         \\
Unique3D     & 50k           & 0.217            & 0.654          & \second{22.243} & \best{0.900}     & \third{0.132}  \\
Trellis lmage-to-3D & 500K     &\best{0.125}      & \best{0.843}   & \third{19.124}  & \second{0.899}   & \second{0.121}   \\
\textbf{Ours}        & 26K            &\second{0.128}    & \second{0.831} & \best{23.223}   & \third{0.898}    & \best{0.119} \\
\hline
Ours w/o Joint Training  & 26K &{0.179}           & {0.719}        & {19.241}        & {0.832}          & {0.141}        \\
Ours (Single-LoRA) & 26K    & 0.172            & 0.691          & 19.122          & 0.811            & 0.149          \\
\bottomrule
\end{tabular}
}
\caption{
Quantitative comparison for \textbf{image-to-3D generation.} We compare our method with existing approaches across geometric (CD, FS) and perceptual (PSNR, SSIM, LPIPS) metrics. Our method achieves strong performance, while using significantly less training data (26K). Notably, \textit{CraftsMan} only reports geometric results (CD and FS) and does not model appearance, hence missing values in PSNR, SSIM, and LPIPS.
}
\label{tab:quantitative_comparison}
\end{table}

\begin{figure}
    \centering
    \includegraphics[trim=20 0 20 0,width=0.9\linewidth]{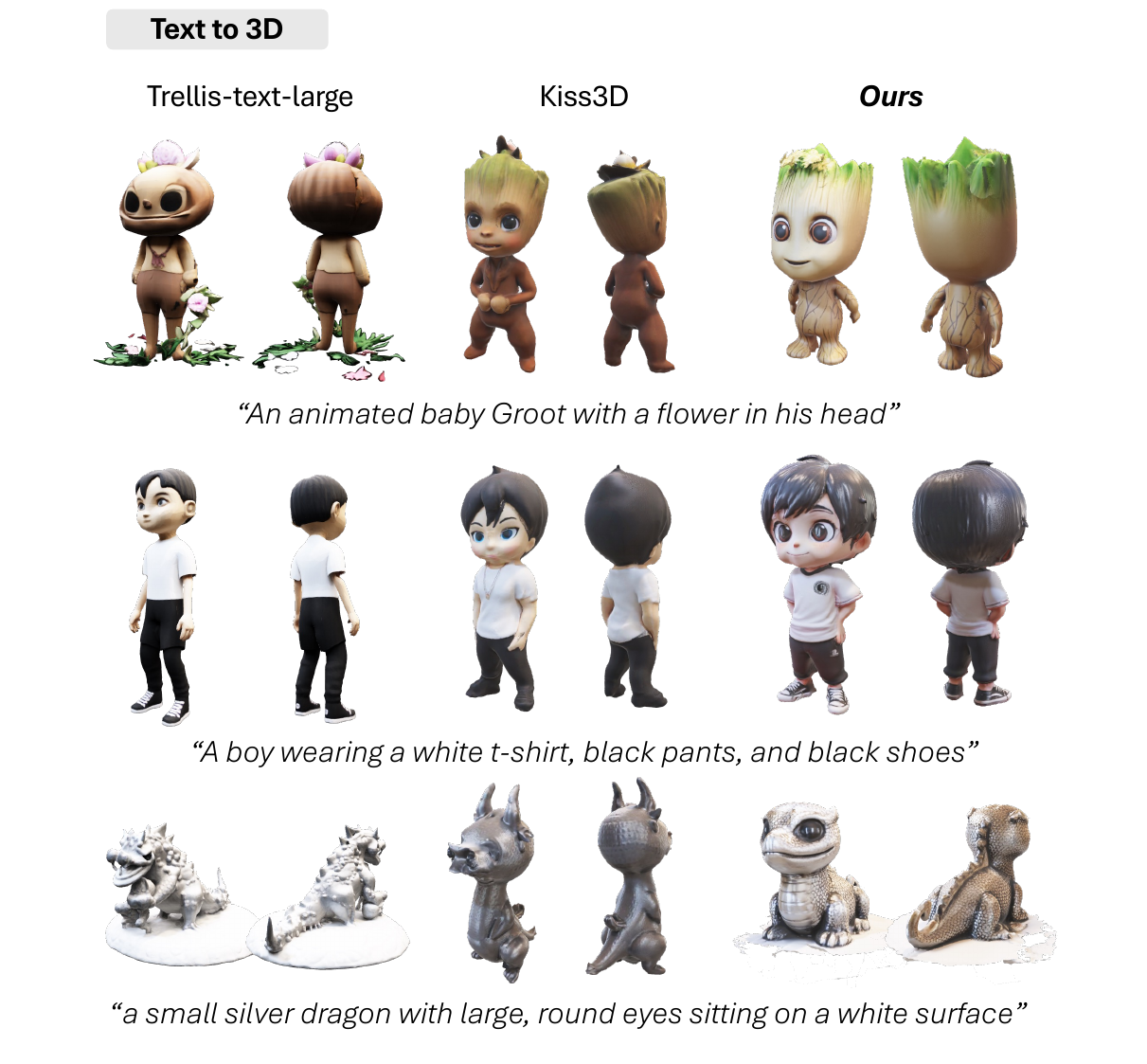}
    \caption{Qualitative comparison on \textbf{text-to-3D generation}. Please zoom in for detail.}
    \vspace{-1em}
    \label{fig:txt-to-3D}
\end{figure}

\noindent\textbf{Text-to-3D generation.}
As shown in Table~\ref{tab:clip_quality_aesthetic}, our method achieves the best overall performance across all three metrics: CLIP score, quality, and aesthetic scores. In particular, we outperform other approaches by a clear margin in terms of text-image alignment and perceptual quality. These trends are also evident in the qualitative results shown in Figure~\ref{fig:txt-to-3D}. Compared with Trellis-text-large and Kiss3DGen~\cite{lin2025kiss3dgen}, our method generates 3D assets that more faithfully capture both the semantic content and fine-grained visual attributes described in the prompts. For example, in the “baby Groot” example, our model produces a character with more pronounced baby-like features—such as rounder eyes, a softer expression, and a more stylized, animated appearance—better matching the intended description. We attribute this to our unified 2.5D representation and effective adaptation of 2D diffusion priors, which together enable precise control over both appearance and structure. These results demonstrate the effectiveness of our approach in generating semantically grounded and visually appealing 3D assets from diverse natural language inputs.

\begin{table}[h]
\centering
\resizebox{0.95\linewidth}{!}{
\begin{tabular}{ccccc}
\toprule
Method   & Dataset Size    & CLIP$\uparrow$ & Quality$\uparrow$ & Aesthetic$\uparrow$ \\
\midrule
3DTopia   &  320K          & 0.693              & 2.145           & \third{1.538}             \\
Hunyuan3D-1 & N/A         & \second{0.792}              & \second{2.517}           & 1.504             \\
Trellis Text-to-3D  & 500K        & \third{0.782}              & \third{2.468}           & \second{1.831}             \\
Ours   & 26K              & \best{0.840}              & \best{2.972}           & \best{1.921}             \\
\bottomrule
\end{tabular}
}
\caption{Quantitative comparison of \textbf{text-to-3D generation} results across CLIP score, perceptual quality, and aesthetic rating. Our method outperforms all baselines on all three metrics.}
\label{tab:clip_quality_aesthetic}
\end{table}

\begin{figure}[t]
    \centering
    \includegraphics[width=0.9\linewidth]{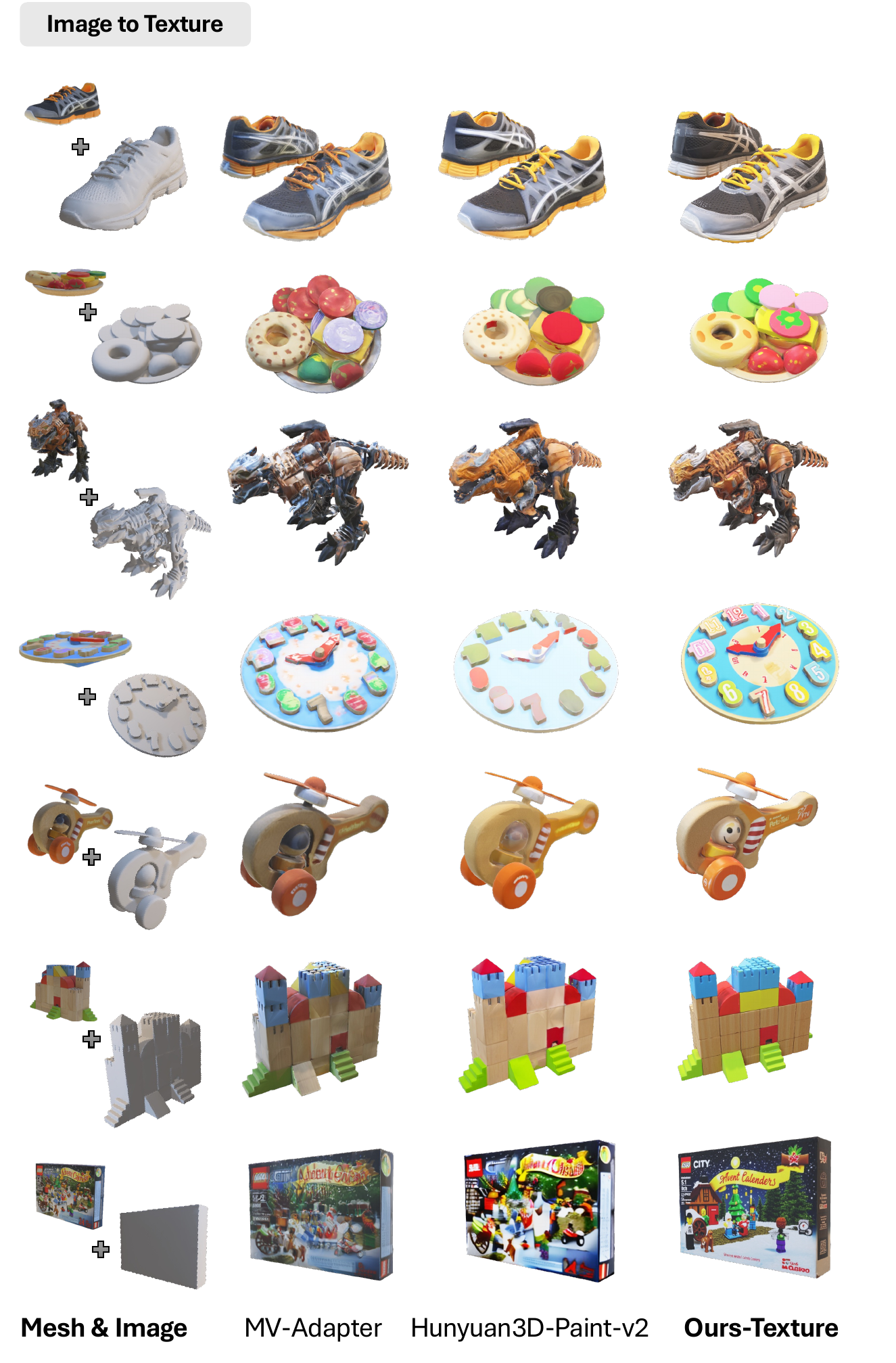}
    \vspace{-1em}
    \caption{Qualitative comparison of \textbf{image and geometry conditioned texture generation}. Our model perform the best in terms of texture fidelity. Please zoom in for detail.}
    \vspace{-1.0em}
    \label{fig:img-tex-gen-quali}
\end{figure}

\begin{figure}[t]
    \centering
    \includegraphics[width=1.\linewidth]{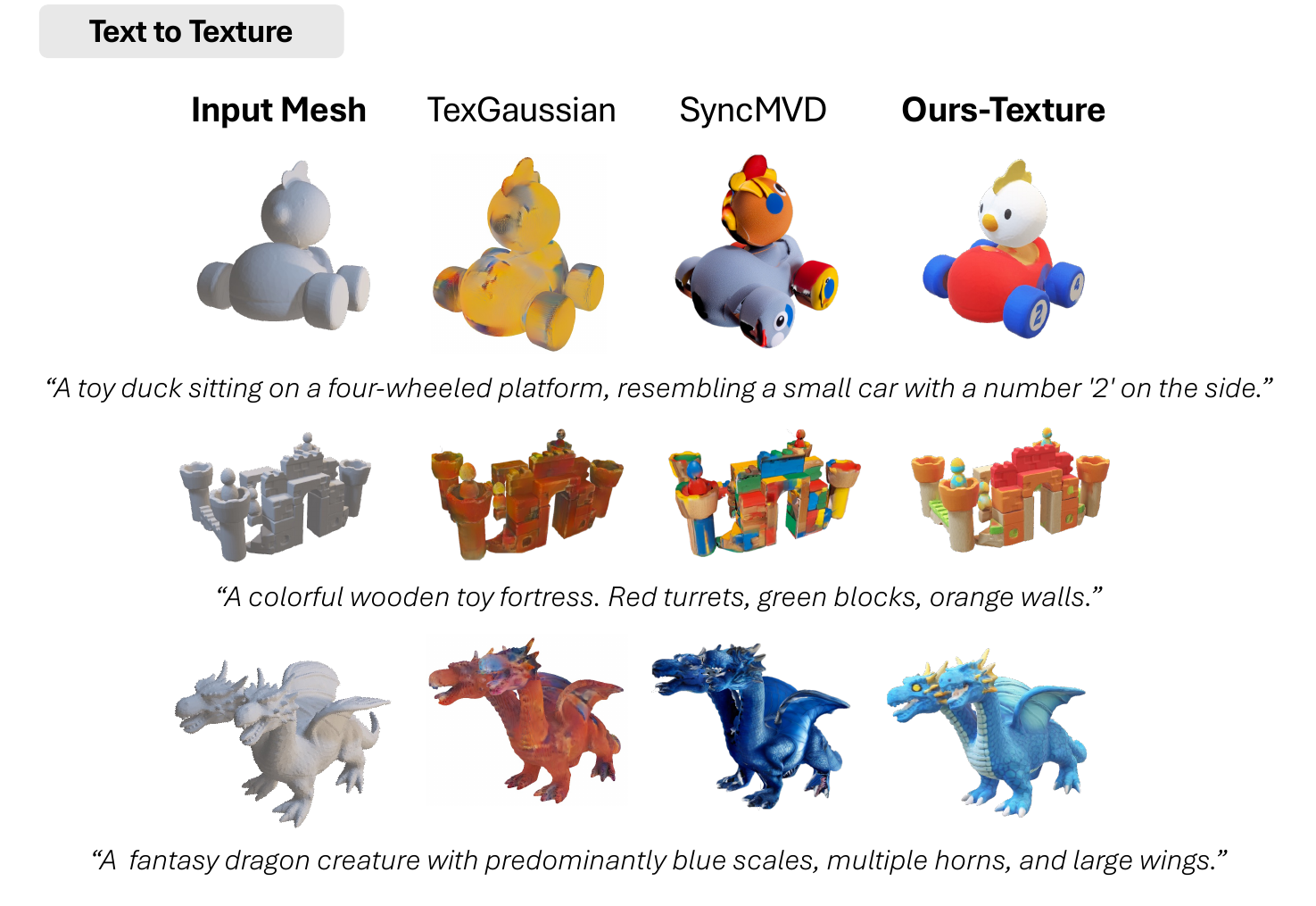}
    \vspace{-1.5em}
    \caption{Qualitative comparison of \textbf{text and geometry conditioned texture generation}. Compare to the baseline methods, our model better interprets the text condition and generates texture coherent with the given geometry conditions. }
    \label{fig:text-tex-gen-quali}
\end{figure}

\noindent\textbf{Texture generation.}
As shown in Table~\ref{tab:texture_text_condition} and Table~\ref{tab:texture_image_condition}, our method consistently outperforms previous approaches under both text and image conditions. For the image condition, our model achieves the best PSNR and SSIM, along with the lowest LPIPS, indicating superior texture fidelity and perceptual similarity. Qualitative results in Figure~\ref{fig:img-tex-gen-quali} further support this: our method preserves fine details such as clear logos and legible text on toy boxes, and produces more accurate material and color rendering compared to baselines like MV-Adapter and Hunyuan3D-Paint-v2. For the text condition, our method achieves higher CLIP scores, perceptual quality, and aesthetic ratings than TexGaussian and SyncMVD. As shown in Figure~\ref{fig:text-tex-gen-quali}, our textures align more closely with the input prompts—for example, correctly capturing distinct color elements like red turrets and green blocks on the toy fortress, or producing vibrant blue scales and detailed features on the fantasy dragon. In contrast, baseline methods often exhibit color bleeding, texture artifacts, or incomplete interpretation of text cues. We attribute these improvements to our unified 2.5D representation, which tightly couples geometry and appearance by encoding multiview RGB, normals, and coordinates. This design ensures better consistency and semantic alignment during texture generation, regardless of the input modality.

\begin{table}[h]
\centering
\resizebox{0.8\linewidth}{!}{
\begin{tabular}{ccccc}
\hline
Method & CLIP-score$\uparrow$ & Quality$\uparrow$ & Aesthetic$\uparrow$ \\
\hline
SyncMVD       & \second{0.798} & \second{3.252} & \second{2.180} \\
TexGaussian  & \third{0.686}  & \third{2.444}  & \third{1.854} \\
Ours          & \best{0.803}   & \best{3.781}   & \best{2.224} \\
\hline
\end{tabular}
}
\caption{Qualitative comparison of \textbf{text-to-texture generation} results.}
\vspace{-2.7em}
\label{tab:texture_text_condition}
\end{table}

\begin{table}[h]
\centering
\resizebox{0.7\linewidth}{!}{
\begin{tabular}{cccc}
\hline
Method & PSNR$\uparrow$ & SSIM$\uparrow$ & LPIPS$\downarrow$ \\
\hline
Hunyuan2.0   & \third{19.688} & \third{0.857} & \second{0.141} \\
MV-adapter   & \second{22.404} & \second{0.871} & \third{0.145} \\
Ours         & \best{24.123} & \best{0.881} & \best{0.124} \\
\hline
\end{tabular}
}
\caption{Qualitative comparison of \textbf{image-to-texture generation} results.}
\label{tab:texture_image_condition}
\end{table}

\subsection{Ablation study}
In this section, we aim to exanimate the effectiveness of our proposed architecture and our training strategy for text and image to 3D generation.

\begin{figure}
    \centering
    \includegraphics[width=0.9\linewidth]{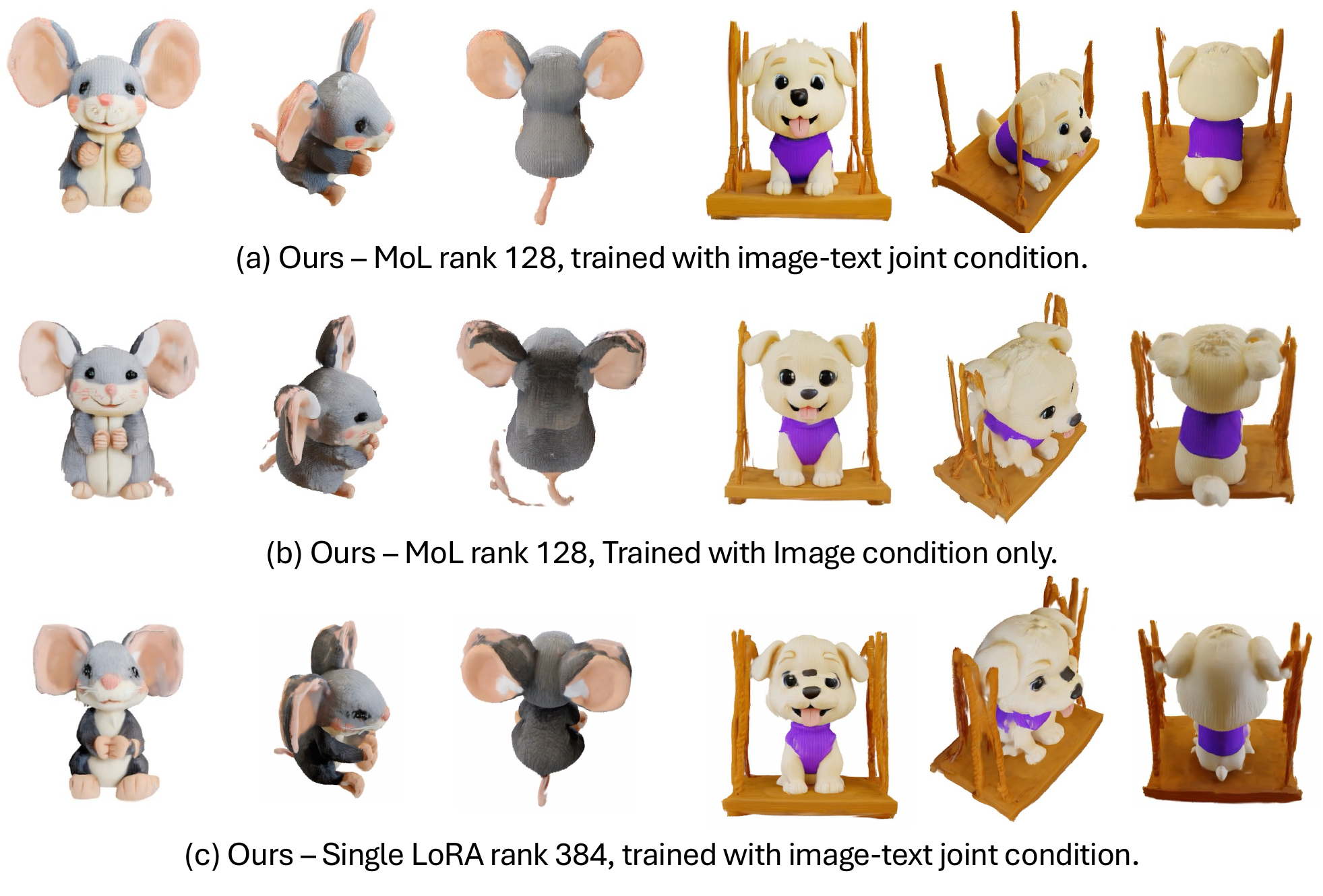}
    \vspace{-1em}
    \caption{Qualitative comparison of image-to-3D generation results (decoded from 2.5D latent to 3D Gaussian Splats) of our models trained with different strategy. The base model trained with Mixture-of-LoRA and image-text joint condition perform significantly better in producing results with better fidelity and 3D coherence. Please zoom in for detail.}
    \vspace{-1.5em}
    \label{fig:ablation}
\end{figure}

\noindent \textbf{Joint condition training.}
we train an image conditioned and a text-or-image conditioned model by randomly disable the image condition during the training.
As shown in Table~\ref{tab:quantitative_comparison}, the jointly trained model performs better in image-conditioned 3D generation. Qualitative comparisons in Figure~\ref{fig:ablation}(b) further show that it produces more coherent and detailed structures, particularly in complex regions. This improvement suggests that exposing the model to diverse conditioning signals during training helps it learn more robust and generalized feature representations, which in turn enhances structural consistency and overall generation quality.

\noindent \textbf{Large rank LoRA v.s. MoL.}
\label{sec:lora-vs-mol}
To evaluate the effectiveness of our propose MoL for 2.5D generation, we setup a comparison between the models trained with single LoRA adapter and MoL with equivalent parameters. 
Specifically, we set the rank of the single LoRA adapter to 384. For MoL, we set rgb, normal and coordinate LoRA adapters with rank of 128, 128 and 128 respectively. Both models trained from scratch with identical training settings.
In Tab.~\ref{tab:quantitative_comparison}, we measure the scores of both models and the MoL model outperform the other with a significant margin. 
And in Fig.~\ref{fig:ablation} (c), we compare both models qualitatively, that our proposed model with MoL deliver significantly better results.

\section{Limitation}
Extensive experiments demonstrate that our approach surpasses existing methods in terms of color-geometry coherence and detail richness for both image- and text-to-3D asset generation, as well as geometry-conditioned texture generation. Furthermore, our method achieves comparable performance to state-of-the-art 3D generative models in terms of geometry coherence and fidelity. 
Nevertheless, there remains room for improvement within our framework, particularly concerning the quality of 3D reconstruction from the 2.5D latents.  The full potential of 2D-3D hybrid representations has yet to be explored and we will leave this problem for future study.

\section{Conclusion}
In this paper, we introduce a novel framework for 3D object generation and geometry-conditioned texture generation by leveraging the capabilities of 2D diffusion models through the concept of 2.5D latents. These latents function as image-like structures that encapsulate the color, surface, and coordinate information of 3D objects, allowing them to be seamlessly projected into structured 3D representations. To facilitate the efficient generation of 2.5D latents, we have developed an innovative Mixture-of-LoRA architecture. Furthermore, a 3D refiner and decoder architecture is employed to effectively reconstruct 3D objects from these latents.

{
    \small
    \bibliographystyle{ieeenat_fullname}
    \bibliography{main}

\begin{thebibliography}{47}
\providecommand{\natexlab}[1]{#1}
\providecommand{\url}[1]{\texttt{#1}}
\expandafter\ifx\csname urlstyle\endcsname\relax
  \providecommand{\doi}[1]{doi: #1}\else
  \providecommand{\doi}{doi: \begingroup \urlstyle{rm}\Url}\fi

\bibitem[Albergo and Vanden-Eijnden(2022)]{albergo2022building}
Michael~S. Albergo and Eric Vanden-Eijnden.
\newblock Building normalizing flows with stochastic interpolants, 2022.

\bibitem[BlackForestLabs(2024)]{blackforest2024flux}
BlackForestLabs.
\newblock Flux.1 model family.
\newblock 2024.

\bibitem[Blattmann et~al.(2023{\natexlab{a}})Blattmann, Dockhorn, Kulal, Mendelevitch, Kilian, Lorenz, Levi, English, Voleti, Letts, et~al.]{blattmann2023stable}
Andreas Blattmann, Tim Dockhorn, Sumith Kulal, Daniel Mendelevitch, Maciej Kilian, Dominik Lorenz, Yam Levi, Zion English, Vikram Voleti, Adam Letts, et~al.
\newblock Stable video diffusion: Scaling latent video diffusion models to large datasets.
\newblock \emph{arXiv preprint arXiv:2311.15127}, 2023{\natexlab{a}}.

\bibitem[Blattmann et~al.(2023{\natexlab{b}})Blattmann, Rombach, Ling, Dockhorn, Kim, Fidler, and Kreis]{blattmann2023align}
Andreas Blattmann, Robin Rombach, Huan Ling, Tim Dockhorn, Seung~Wook Kim, Sanja Fidler, and Karsten Kreis.
\newblock Align your latents: High-resolution video synthesis with latent diffusion models.
\newblock In \emph{Proceedings of the IEEE/CVF conference on computer vision and pattern recognition}, pages 22563--22575, 2023{\natexlab{b}}.

\bibitem[Chen et~al.(2024)Chen, Shen, Liu, Shi, Zhou, Lin, Gu, Su, Wetzstein, and Guibas]{chen20243d}
Hansheng Chen, Bokui Shen, Yulin Liu, Ruoxi Shi, Linqi Zhou, Connor~Z Lin, Jiayuan Gu, Hao Su, Gordon Wetzstein, and Leonidas Guibas.
\newblock 3d-adapter: Geometry-consistent multi-view diffusion for high-quality 3d generation.
\newblock \emph{arXiv preprint arXiv:2410.18974}, 2024.

\bibitem[Downs et~al.(2022)Downs, Francis, Koenig, Kinman, Hickman, Reymann, McHugh, and Vanhoucke]{downs2022google}
Laura Downs, Anthony Francis, Nate Koenig, Brandon Kinman, Ryan Hickman, Krista Reymann, Thomas~B McHugh, and Vincent Vanhoucke.
\newblock Google scanned objects: A high-quality dataset of 3d scanned household items.
\newblock In \emph{ICRA}, 2022.

\bibitem[Esser et~al.(2024)Esser, Kulal, Blattmann, Entezari, M{\"u}ller, Saini, Levi, Lorenz, Sauer, Boesel, et~al.]{esser2024scaling}
Patrick Esser, Sumith Kulal, Andreas Blattmann, Rahim Entezari, Jonas M{\"u}ller, Harry Saini, Yam Levi, Dominik Lorenz, Axel Sauer, Frederic Boesel, et~al.
\newblock Scaling rectified flow transformers for high-resolution image synthesis.
\newblock In \emph{Forty-first international conference on machine learning}, 2024.

\bibitem[He et~al.(2024{\natexlab{a}})He, Li, Hu, Shen, Cai, Qiu, and Chen]{he2024disenvisioner}
Jing He, Haodong Li, Yongzhe Hu, Guibao Shen, Yingjie Cai, Weichao Qiu, and Ying-Cong Chen.
\newblock Disenvisioner: Disentangled and enriched visual prompt for customized image generation.
\newblock \emph{arXiv preprint arXiv:2410.02067}, 2024{\natexlab{a}}.

\bibitem[He et~al.(2024{\natexlab{b}})He, Li, Yin, Liang, Li, Zhou, Zhang, Liu, and Chen]{he2024lotus}
Jing He, Haodong Li, Wei Yin, Yixun Liang, Leheng Li, Kaiqiang Zhou, Hongbo Zhang, Bingbing Liu, and Ying-Cong Chen.
\newblock Lotus: Diffusion-based visual foundation model for high-quality dense prediction.
\newblock \emph{arXiv preprint arXiv:2409.18124}, 2024{\natexlab{b}}.

\bibitem[Hessel et~al.(2021)Hessel, Holtzman, Forbes, Bras, and Choi]{hessel2021clipscore}
Jack Hessel, Ari Holtzman, Maxwell Forbes, Ronan~Le Bras, and Yejin Choi.
\newblock Clipscore: A reference-free evaluation metric for image captioning.
\newblock In \emph{EMNLP}, 2021.

\bibitem[Ho et~al.(2020)Ho, Jain, and Abbeel]{ho2020denoising}
Jonathan Ho, Ajay Jain, and Pieter Abbeel.
\newblock Denoising diffusion probabilistic models.
\newblock \emph{Advances in neural information processing systems}, 33:\penalty0 6840--6851, 2020.

\bibitem[Ho et~al.(2022{\natexlab{a}})Ho, Chan, Saharia, Whang, Gao, Gritsenko, Kingma, Poole, Norouzi, Fleet, and Salimans]{ho2022imagen}
Jonathan Ho, William Chan, Chitwan Saharia, Jay Whang, Ruiqi Gao, Alexey Gritsenko, Diederik~P. Kingma, Ben Poole, Mohammad Norouzi, David~J. Fleet, and Tim Salimans.
\newblock Imagen video: High definition video generation with diffusion models, 2022{\natexlab{a}}.

\bibitem[Ho et~al.(2022{\natexlab{b}})Ho, Salimans, Gritsenko, Chan, Norouzi, and Fleet]{ho2022video}
Jonathan Ho, Tim Salimans, Alexey Gritsenko, William Chan, Mohammad Norouzi, and David~J Fleet.
\newblock Video diffusion models.
\newblock \emph{Advances in Neural Information Processing Systems}, 35:\penalty0 8633--8646, 2022{\natexlab{b}}.

\bibitem[Hong et~al.(2024)Hong, Tang, Cao, Shi, Wu, Chen, Wang, Pan, Lin, and Liu]{hong20243dtopia}
Fangzhou Hong, Jiaxiang Tang, Ziang Cao, Min Shi, Tong Wu, Zhaoxi Chen, Tengfei Wang, Liang Pan, Dahua Lin, and Ziwei Liu.
\newblock 3dtopia: Large text-to-3d generation model with hybrid diffusion priors.
\newblock \emph{arXiv preprint arXiv:2403.02234}, 2024.

\bibitem[Hu et~al.(2022)Hu, Shen, Wallis, Allen-Zhu, Li, Wang, Wang, and Chen]{hu2022lora}
Edward~J Hu, Yelong Shen, Phillip Wallis, Zeyuan Allen-Zhu, Yuanzhi Li, Shean Wang, Lu Wang, and Weizhu Chen.
\newblock Lo{RA}: Low-rank adaptation of large language models.
\newblock In \emph{International Conference on Learning Representations}, 2022.

\bibitem[Huang et~al.(2024)Huang, Guo, Wang, Yi, Ma, Cao, and Sheng]{huang2024mvadapter}
Zehuan Huang, Yuanchen Guo, Haoran Wang, Ran Yi, Lizhuang Ma, Yan-Pei Cao, and Lu Sheng.
\newblock Mv-adapter: Multi-view consistent image generation made easy.
\newblock \emph{arXiv}, 2024.

\bibitem[Karras et~al.(2022)Karras, Aittala, Aila, and Laine]{Karras2022ElucidatingTD}
Tero Karras, Miika Aittala, Timo Aila, and Samuli Laine.
\newblock Elucidating the design space of diffusion-based generative models.
\newblock \emph{ArXiv}, abs/2206.00364, 2022.

\bibitem[Kerbl et~al.(2023)Kerbl, Kopanas, Leimk{\"u}hler, and Drettakis]{kerbl20233d}
Bernhard Kerbl, Georgios Kopanas, Thomas Leimk{\"u}hler, and George Drettakis.
\newblock 3d gaussian splatting for real-time radiance field rendering.
\newblock \emph{ACM Trans. Graph.}, 42\penalty0 (4):\penalty0 139--1, 2023.

\bibitem[Li et~al.(2024)Li, Liu, Chen, Liang, Chen, Tan, and Long]{li2024craftsman}
Weiyu Li, Jiarui Liu, Rui Chen, Yixun Liang, Xuelin Chen, Ping Tan, and Xiaoxiao Long.
\newblock Craftsman: High-fidelity mesh generation with 3d native generation and interactive geometry refiner.
\newblock \emph{arXiv preprint arXiv:2405.14979}, 2024.

\bibitem[Li et~al.(2025)Li, Zou, Liu, Wang, Liang, Yu, Liu, Guo, Liang, Ouyang, et~al.]{li2025triposg}
Yangguang Li, Zi-Xin Zou, Zexiang Liu, Dehu Wang, Yuan Liang, Zhipeng Yu, Xingchao Liu, Yuan-Chen Guo, Ding Liang, Wanli Ouyang, et~al.
\newblock Triposg: High-fidelity 3d shape synthesis using large-scale rectified flow models.
\newblock \emph{arXiv}, 2025.

\bibitem[Lin et~al.(2025)Lin, Yang, Chen, Xu, Yan, Wu, Xu, Xu, Zhang, and Chen]{lin2025kiss3dgen}
Jiantao Lin, Xin Yang, Meixi Chen, Yingjie Xu, Dongyu Yan, Leyi Wu, Xinli Xu, Lie Xu, Shunsi Zhang, and Ying-Cong Chen.
\newblock Kiss3dgen: Repurposing image diffusion models for 3d asset generation.
\newblock \emph{CVPR}, 2025.

\bibitem[Lipman et~al.(2023)Lipman, Chen, Ben-Hamu, Nickel, and Le]{lipman2023flow}
Yaron Lipman, Ricky T.~Q. Chen, Heli Ben-Hamu, Maximilian Nickel, and Matthew Le.
\newblock Flow matching for generative modeling.
\newblock In \emph{The Eleventh International Conference on Learning Representations}, 2023.

\bibitem[Liu et~al.(2022)Liu, Gong, and Liu]{liu2022flow}
Xingchao Liu, Chengyue Gong, and Qiang Liu.
\newblock Flow straight and fast: Learning to generate and transfer data with rectified flow, 2022.

\bibitem[Liu et~al.(2024)Liu, Xie, Liu, and Wong]{liu2024text}
Yuxin Liu, Minshan Xie, Hanyuan Liu, and Tien-Tsin Wong.
\newblock Text-guided texturing by synchronized multi-view diffusion.
\newblock In \emph{SIGGRAPH Asia 2024 Conference Papers}, 2024.

\bibitem[Liu et~al.(2023)Liu, Feng, Black, Nowrouzezahrai, Paull, and Liu]{liu2023meshdiffusion}
Zhen Liu, Yao Feng, Michael~J Black, Derek Nowrouzezahrai, Liam Paull, and Weiyang Liu.
\newblock Meshdiffusion: Score-based generative 3d mesh modeling.
\newblock \emph{arXiv preprint arXiv:2303.08133}, 2023.

\bibitem[Long et~al.(2023)Long, Guo, Lin, Liu, Dou, Liu, Ma, Zhang, Habermann, Theobalt, et~al.]{long2023wonder3d}
Xiaoxiao Long, Yuan-Chen Guo, Cheng Lin, Yuan Liu, Zhiyang Dou, Lingjie Liu, Yuexin Ma, Song-Hai Zhang, Marc Habermann, Christian Theobalt, et~al.
\newblock Wonder3d: Single image to 3d using cross-domain diffusion.
\newblock \emph{arXiv}, 2023.

\bibitem[Lu et~al.(2024)Lu, Zhang, Li, Fang, McKinnon, Tsin, Quan, Cao, and Yao]{lu2024direct2}
Yuanxun Lu, Jingyang Zhang, Shiwei Li, Tian Fang, David McKinnon, Yanghai Tsin, Long Quan, Xun Cao, and Yao Yao.
\newblock Direct2. 5: Diverse text-to-3d generation via multi-view 2.5 d diffusion.
\newblock In \emph{Proceedings of the IEEE/CVF Conference on Computer Vision and Pattern Recognition}, pages 8744--8753, 2024.

\bibitem[Miao et~al.(2024)Miao, Agrawal, Zhang, Semeraro, Cavallo, Gu, and Toshev]{miao2024dsplats}
Kevin Miao, Harsh Agrawal, Qihang Zhang, Federico Semeraro, Marco Cavallo, Jiatao Gu, and Alexander Toshev.
\newblock Dsplats: 3d generation by denoising splats-based multiview diffusion models.
\newblock \emph{arXiv preprint arXiv:2412.09648}, 2024.

\bibitem[Qin et~al.(2025)Qin, Chen, Kong, Lu, and Zhu]{qin2025distilling}
Hao Qin, Luyuan Chen, Ming Kong, Mengxu Lu, and Qiang Zhu.
\newblock Distilling multi-view diffusion models into 3d generators.
\newblock \emph{arXiv preprint arXiv:2504.00457}, 2025.

\bibitem[Ramesh et~al.(2021)Ramesh, Pavlov, Goh, Gray, Voss, Radford, Chen, and Sutskever]{dalle}
Aditya Ramesh, Mikhail Pavlov, Gabriel Goh, Scott Gray, Chelsea Voss, Alec Radford, Mark Chen, and Ilya Sutskever.
\newblock Zero-shot text-to-image generation.
\newblock In \emph{International Conference on Machine Learning}, pages 8821--8831. PMLR, 2021.

\bibitem[Rombach et~al.(2022)Rombach, Blattmann, Lorenz, Esser, and Ommer]{rombach2022high}
Robin Rombach, Andreas Blattmann, Dominik Lorenz, Patrick Esser, and Bj{\"o}rn Ommer.
\newblock High-resolution image synthesis with latent diffusion models.
\newblock In \emph{Proceedings of the IEEE/CVF conference on computer vision and pattern recognition}, pages 10684--10695, 2022.

\bibitem[Saharia et~al.(2022)Saharia, Chan, Saxena, Li, Whang, Denton, Ghasemipour, Gontijo~Lopes, Karagol~Ayan, Salimans, et~al.]{imagen}
Chitwan Saharia, William Chan, Saurabh Saxena, Lala Li, Jay Whang, Emily~L Denton, Kamyar Ghasemipour, Raphael Gontijo~Lopes, Burcu Karagol~Ayan, Tim Salimans, et~al.
\newblock Photorealistic text-to-image diffusion models with deep language understanding.
\newblock \emph{Advances in Neural Information Processing Systems}, 35:\penalty0 36479--36494, 2022.

\bibitem[Shen et~al.(2023)Shen, Munkberg, Hasselgren, Yin, Wang, Chen, Gojcic, Fidler, Sharp, and Gao]{shen2023flexible}
Tianchang Shen, Jacob Munkberg, Jon Hasselgren, Kangxue Yin, Zian Wang, Wenzheng Chen, Zan Gojcic, Sanja Fidler, Nicholas Sharp, and Jun Gao.
\newblock Flexible isosurface extraction for gradient-based mesh optimization.
\newblock \emph{ACM Transactions on Graphics (TOG)}, 42\penalty0 (4):\penalty0 1--16, 2023.

\bibitem[Shi et~al.(2023)Shi, Wang, Ye, Long, Li, and Yang]{shi2023mvdream}
Yichun Shi, Peng Wang, Jianglong Ye, Mai Long, Kejie Li, and Xiao Yang.
\newblock Mvdream: Multi-view diffusion for 3d generation.
\newblock \emph{arXiv preprint arXiv:2308.16512}, 2023.

\bibitem[Sohl-Dickstein et~al.(2015)Sohl-Dickstein, Weiss, Maheswaranathan, and Ganguli]{SohlDickstein2015DeepUL}
Jascha~Narain Sohl-Dickstein, Eric~A. Weiss, Niru Maheswaranathan, and Surya Ganguli.
\newblock Deep unsupervised learning using nonequilibrium thermodynamics.
\newblock \emph{ArXiv}, abs/1503.03585, 2015.

\bibitem[Team(2024)]{yang2024hunyuan3d}
Tencent~Hunyuan3D Team.
\newblock Hunyuan3d 1.0: A unified framework for text-to-3d and image-to-3d generation, 2024.

\bibitem[Team(2025)]{hunyuan3d22025tencent}
Tencent~Hunyuan3D Team.
\newblock Hunyuan3d 2.0: Scaling diffusion models for high resolution textured 3d assets generation, 2025.

\bibitem[Wang et~al.(2024)Wang, Wang, Chen, Xiang, Chen, Yu, Li, Su, and Zhu]{wang2024crm}
Zhengyi Wang, Yikai Wang, Yifei Chen, Chendong Xiang, Shuo Chen, Dajiang Yu, Chongxuan Li, Hang Su, and Jun Zhu.
\newblock Crm: Single image to 3d textured mesh with convolutional reconstruction model.
\newblock \emph{arXiv}, 2024.

\bibitem[Wu et~al.(2023)Wu, Zhang, Zhang, Chen, Li, Liao, Wang, Zhang, Sun, Yan, Min, Zhai, and Lin]{wu2023qalign}
Haoning Wu, Zicheng Zhang, Weixia Zhang, Chaofeng Chen, Chunyi Li, Liang Liao, Annan Wang, Erli Zhang, Wenxiu Sun, Qiong Yan, Xiongkuo Min, Guangtai Zhai, and Weisi Lin.
\newblock Q-align: Teaching lmms for visual scoring via discrete text-defined levels.
\newblock \emph{arXiv}, 2023.

\bibitem[Wu et~al.(2024)Wu, Liu, Cai, Yan, Wang, Hu, Duan, and Ma]{wu2024unique3d}
Kailu Wu, Fangfu Liu, Zhihan Cai, Runjie Yan, Hanyang Wang, Yating Hu, Yueqi Duan, and Kaisheng Ma.
\newblock Unique3d: High-quality and efficient 3d mesh generation from a single image, 2024.

\bibitem[Xiang et~al.(2024)Xiang, Lv, Xu, Deng, Wang, Zhang, Chen, Tong, and Yang]{xiang2024structured}
Jianfeng Xiang, Zelong Lv, Sicheng Xu, Yu Deng, Ruicheng Wang, Bowen Zhang, Dong Chen, Xin Tong, and Jiaolong Yang.
\newblock Structured 3d latents for scalable and versatile 3d generation.
\newblock \emph{arXiv}, 2024.

\bibitem[Xiao et~al.(2023)Xiao, Wu, Xu, Dai, Hu, Lu, Zeng, Liu, and Yuan]{xiao2023florence}
Bin Xiao, Haiping Wu, Weijian Xu, Xiyang Dai, Houdong Hu, Yumao Lu, Michael Zeng, Ce Liu, and Lu Yuan.
\newblock Florence-2: Advancing a unified representation for a variety of vision tasks.
\newblock \emph{arXiv}, 2023.

\bibitem[Xiong et~al.(2024)Xiong, Liu, Hu, Wu, Wu, Liu, Zhao, Ding, and Lian]{xiong2024texgaussian}
Bojun Xiong, Jialun Liu, Jiakui Hu, Chenming Wu, Jinbo Wu, Xing Liu, Chen Zhao, Errui Ding, and Zhouhui Lian.
\newblock Texgaussian: Generating high-quality pbr material via octree-based 3d gaussian splatting.
\newblock \emph{arXiv}, 2024.

\bibitem[Xu et~al.(2024)Xu, Ge, Lin, Feng, Xu, Zhao, Zhang, and Chen]{xu2024flexgen}
Xinli Xu, Wenhang Ge, Jiantao Lin, Jiawei Feng, Lie Xu, HanFeng Zhao, Shunsi Zhang, and Ying-Cong Chen.
\newblock Flexgen: Flexible multi-view generation from text and image inputs.
\newblock \emph{arXiv preprint arXiv:2410.10745}, 2024.

\bibitem[Zhai et~al.(2023)Zhai, Mustafa, Kolesnikov, and Beyer]{zhai2023sigmoidlosslanguageimage}
Xiaohua Zhai, Basil Mustafa, Alexander Kolesnikov, and Lucas Beyer.
\newblock Sigmoid loss for language image pre-training, 2023.

\bibitem[Zhang et~al.(2024)Zhang, Wang, Zhang, Qiu, Pang, Jiang, Yang, Xu, and Yu]{zhang2024clay}
Longwen Zhang, Ziyu Wang, Qixuan Zhang, Qiwei Qiu, Anqi Pang, Haoran Jiang, Wei Yang, Lan Xu, and Jingyi Yu.
\newblock Clay: A controllable large-scale generative model for creating high-quality 3d assets.
\newblock \emph{ACM Transactions on Graphics (TOG)}, 43\penalty0 (4):\penalty0 1--20, 2024.

\bibitem[Zheng et~al.(2024)Zheng, Pan, Guo, Tong, and Liu]{zheng2024mvd}
Xin-Yang Zheng, Hao Pan, Yu-Xiao Guo, Xin Tong, and Yang Liu.
\newblock Mvd\^{} 2: Efficient multiview 3d reconstruction for multiview diffusion.
\newblock In \emph{ACM SIGGRAPH 2024 Conference Papers}, 2024.

\end{thebibliography}
}

\appendix

\begin{figure*}[t]
    \centering
    \includegraphics[width=0.98\linewidth]{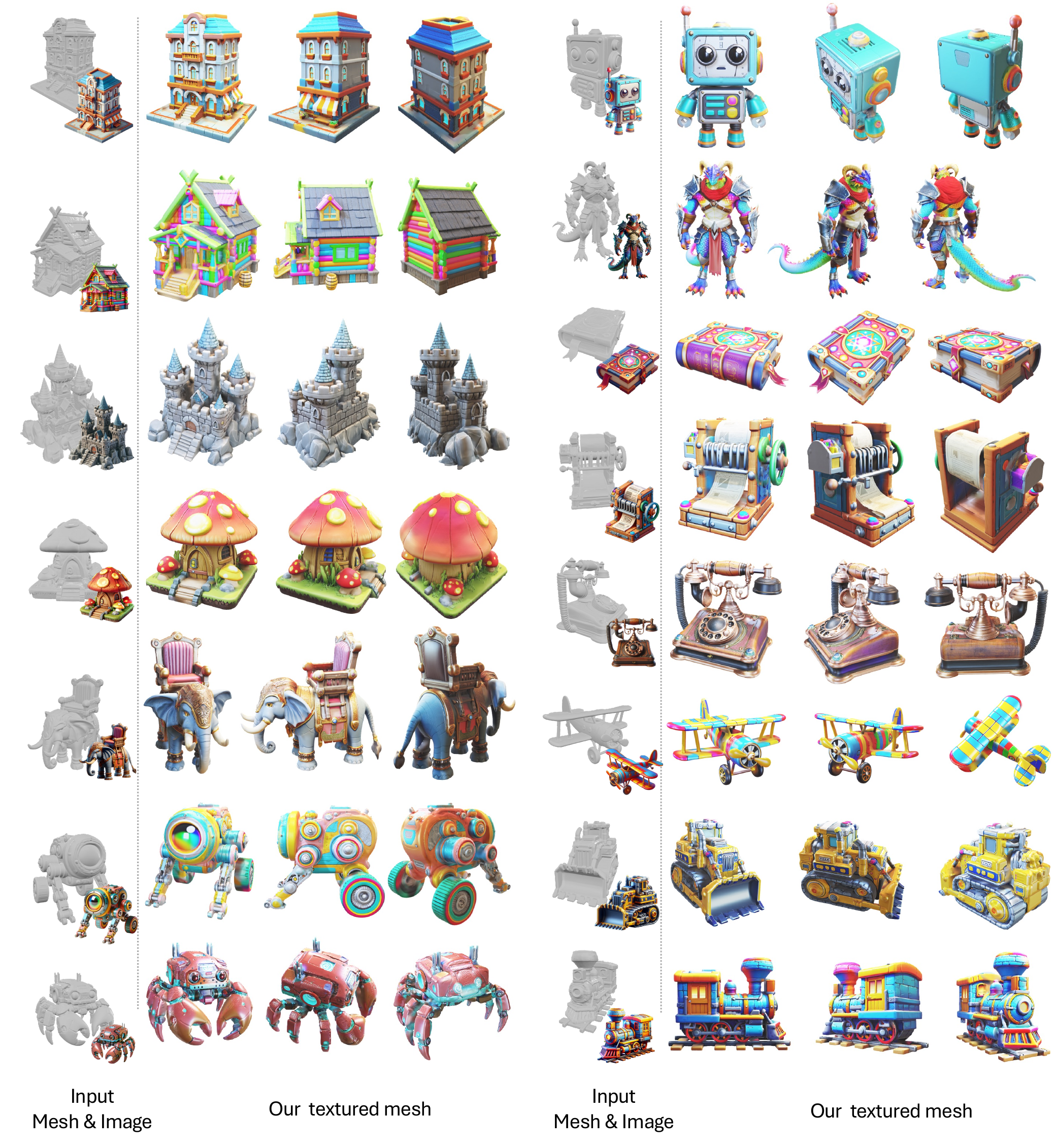}
    \vspace{-1em}
    \caption{Examples of \textbf{image-and-geometry-to-texture} generation with our framework. Please zoom in for detail.}
    \label{fig:i23-gallery}
\end{figure*}

\begin{figure*}[t]
    \centering
    \includegraphics[width=0.98\linewidth]{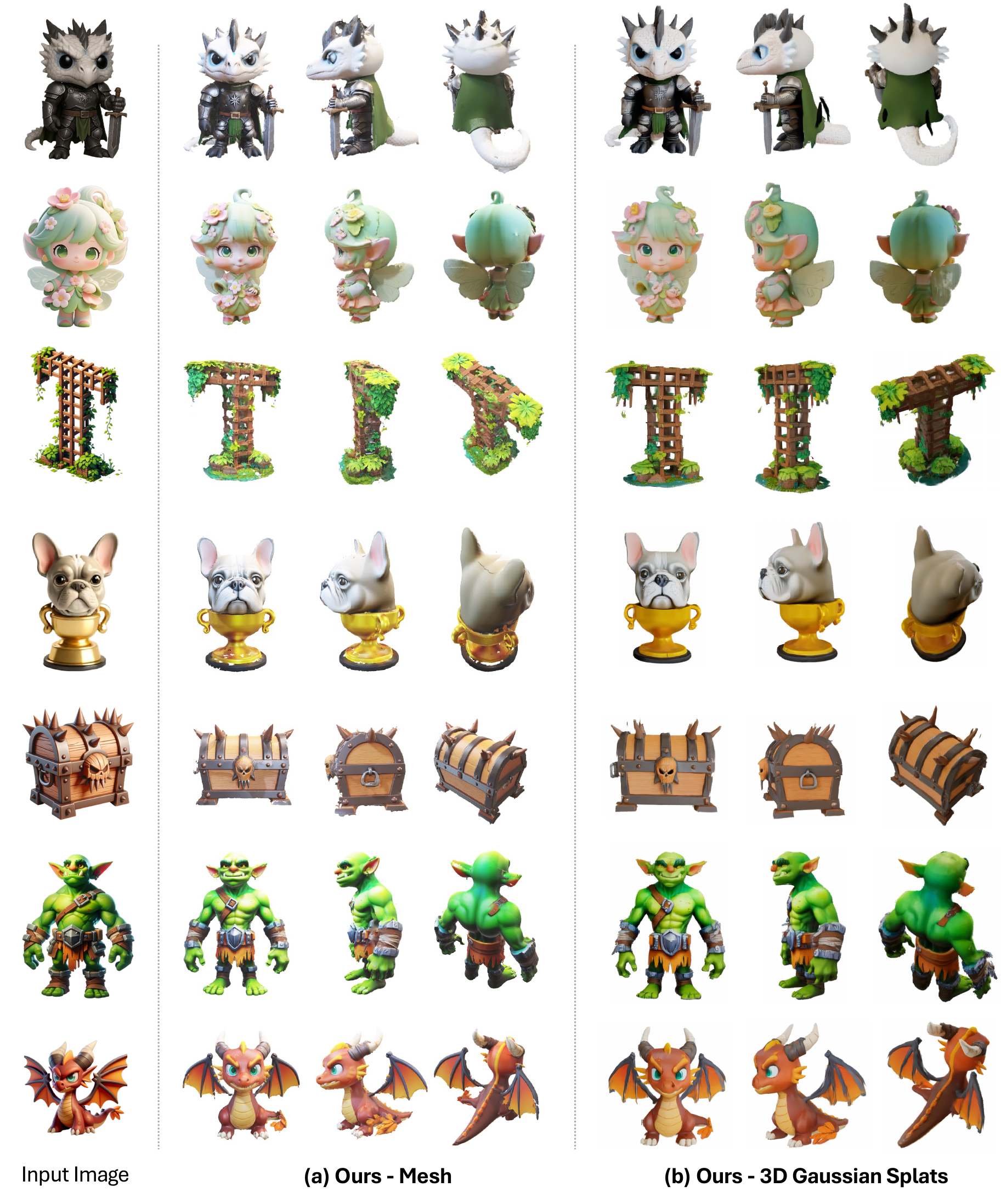}
    \vspace{-1em}
    \caption{Examples of \textbf{image-to-3D} generation with our framework. (a) We apply an extra refinement process to improve the surface and texture detail on the mesh decoder output using ISOMER~\cite{wu2024unique3d}. (b) The 3DGS results are rendered from decoder output directly. Please zoom in for detail.}
    \label{fig:i23-gallery}
\end{figure*}

\section{Model Details}
\subsection{Mixture-of-LoRA}
In our paper, we present a new strategy for fine-tuning a pretrained 2D diffusion model using a Mixture-of-LoRA approach. Notably, in our base model, we introduce three different LoRA adapters, named "General," "Normal," and "Coord." Specifically, all adapters are applied to the "to\_q," "to\_k," "to\_v," "to\_out," and the feedforward layers of the base model. Additionally, we apply only the "General" adapter to the context projection layers, such as "add\_q\_proj," "add\_k\_proj," and "add\_v\_proj." In our experiments, all LoRA layers are set to a rank of 128 except for the single-LoRA model (rank 384).

\subsection{Extended Rotary Positional Embedding}
In our main paper, we propose a modification to the intrinsically integrated rotary positional embedding (RoPE) strategy within the base model, Flux.1-dev, to better preserve its properties when applied to 2.5D images. This modification employs a straightforward approach that enhances positional embedding for multi-modal images.

Specifically, the RoPE in the Flux models is initialized as $p \in \mathbb{R}^3$. In this setup, the first dimension of $p$ is consistently set to zero, while the last two dimensions of $p$ represent the 2D coordinates corresponding to each latent pixel. Our approach treats different modalities within the 2.5D images, such as RGB, normal, and coordinate values, as separate images with identical spatial resolutions. These modalities are assumed to be aligned in the 3D space, which means they are processed collectively during the generation of the 2.5D latents.
To achieve this, for the same spatial position across the RGB, normal, and coordinate latents, we apply identical positional embeddings to the last two dimensions of $p$. Additionally, we introduce constant biases of 0, 32, and 64 in the first dimension of $p$ for RGB, normal, and coordinate latents, respectively.

Our experiments demonstrate that this method provides an effective positional embedding for 2.5D images, achieving good alignment across different modalities.

\subsection{3D Sparse Decoder for 3DGS and Mesh}
To decode our structured 2.5D latents into different 3D output formats, we implement task-specific sparse decoders for both 3D Gaussians (3DGS) and meshes. Both decoders operate over a $64^3$ sparse grid and take as input 48-channel voxel features, obtained by concatenating 16-dimensional latents from RGB, normal, and coordinate branches of the Flux VAE encoder.

\noindent\textbf{3DGS Decoder.}
The 3D Gaussian decoder is designed for high-resolution, high-fidelity reconstruction. It uses 12 Swin-style attention blocks with 16 heads and 512 hidden channels, with a window size of 8 and MLP ratio of 4. The decoder outputs Gaussian attributes including position offsets, colors, opacities, scales, and rotations.

\noindent\textbf{Mesh Decoder.}
The mesh decoder shares the same architectural backbone but uses 12 heads and a wider hidden size of 768. It predicts signed distance values for voxel-centered cubes following the FlexiCubes~\cite{shen2023flexible} representation. Two convolutional upsampling layers are applied post-decoding to increase spatial resolution and extract 0-level isosurfaces as meshes.

\section{Inference Details}
\noindent \textbf{Post generation coordinate correction.}
After generate the 2.5D latents, we aim to project the 2.5D latent features into 3D voxel space for further decoding, which requires to decode the coordinate map from 2.5D latents.
However, we discover that the initial decoded coordinate maps, denoted as $\mathbf{C}$, are usually noisy due to the lossy, uncertain VAE decoding process.
Hence, we introduce a simple pipeline to post process on $\mathbf{C}$ for better accuracy.

First, we project the coordinate maps into point clouds, and apply a bilateral filter on the points to acquire a lower resolution, but confident point cloud $\mathbf{C}'$.
Since we train the model with 2.5D images rendered from 3D objects with known camera parameters, we are able to apply a camera ray regularization on the generated point clouds, that for each point in $\mathbf{C}'$, it must on the ray which pointing from the camera origin to the pixel on the coordinate map image. To ensure the regularization, we project each points in $\mathbf{C}'$ on the ray they supposed to be on.

\begin{figure*}[t]
    \centering
    \includegraphics[width=1.0\linewidth]{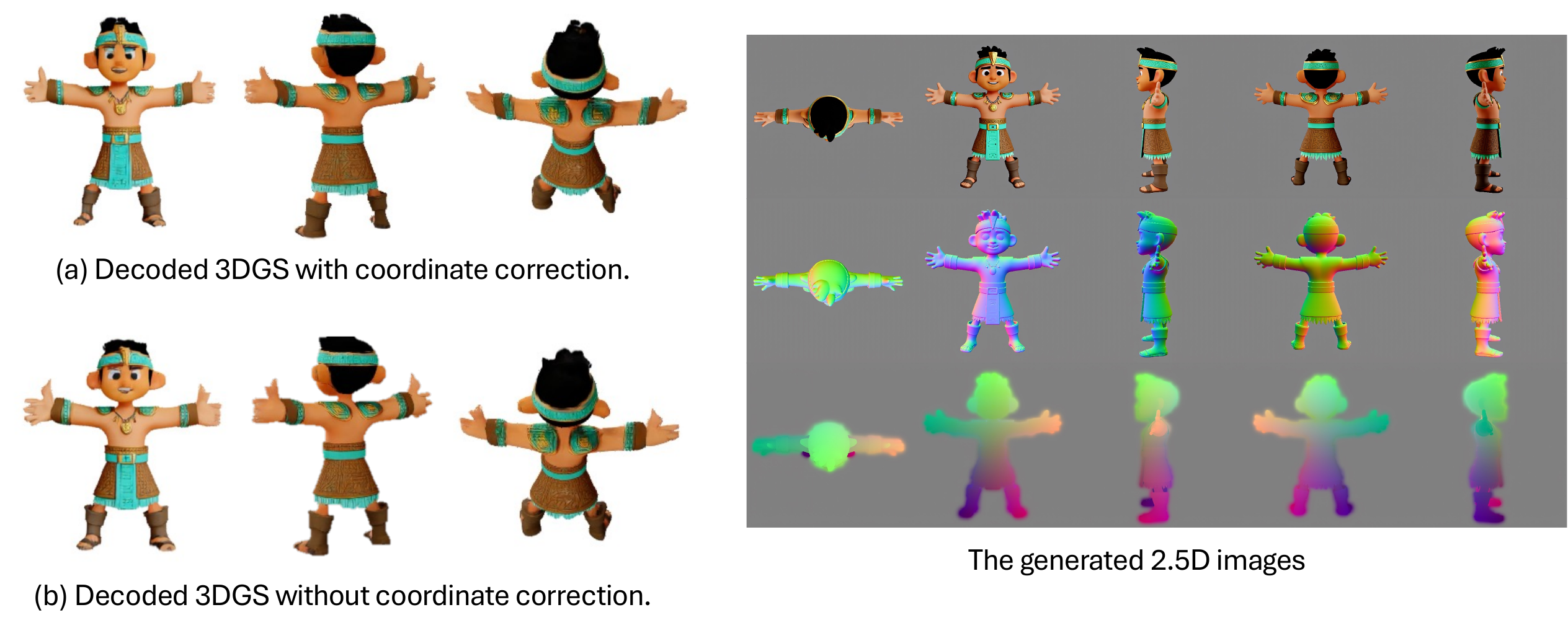}
    \caption{Qualitative comparison between the decoded 3DGS from generated 2.5D images (a) with or (b) without coordinate correction. The result with coordinate correction has better structural alignment with the generated multiview 2.5D images. Please zoom in for details.}
    \label{fig:abla-3d-correct}
\end{figure*}

After the projection, we obtain a new, camera regularized point cloud $\mathbf{C}''$, which we then extract the point projection direction by $\vec{\mathbf{C}} = \mathbf{C}'' - \mathbf{C}$, then applied another spatial bilateral filter on $\vec{\mathbf{C}}$ to remove the noise from the projection.
With the filtered point projection direction $\vec{\mathbf{C}}'$, we produce the final corrected point cloud $\hat{C} = C + \vec{\mathbf{C}}'$. Fig.~\ref{fig:abla-3d-correct} demonstrate the difference between the original and the coordinate corrected results.


\end{document}